%% file: iclr2026_conference.tex
\documentclass{article} 
\usepackage{iclr2026_conference,times}

\input{math_commands.tex}

\usepackage{hyperref}
\usepackage{url}
\input{papel/import}

\title{Beyond Patient Invariance: Learning Cardiac Dynamics via Action-Conditioned JEPAs}

\author{Jose Geraldo Fernandes, Luiz Facury, Pedro Robles Dutenhefner \& Wagner Meira Jr.
\\
Department of Computer Science\\
Universidade Federal de Minas Gerais\\
Belo Horizonte, Brazil \\
\texttt{\{josefernandes,luizfacury,pedroroblesduten,meira\}@dcc.ufmg.br} \\
}

\iclrfinalcopy 
\begin{document}

\maketitle

\input{papel/abstract}

\input{papel/introduction}
\input{papel/related}
\input{papel/method}
\input{papel/experiments}
\input{papel/conclusion}

\bibliography{papel/refs}
\bibliographystyle{iclr2026_conference}

\input{papel/appendix}

\end{document}

%% file: math_commands.tex
\usepackage{amsmath,amsfonts,bm}

\def\eqref#1{equation~\ref{#1}}

\def\1{\bm{1}}

\def\va{{\bm{a}}}

\def\vh{{\bm{h}}}

\def\vy{{\bm{y}}}

\def\mH{{\bm{H}}}
\def\mI{{\bm{I}}}

\def\mX{{\bm{X}}}

\DeclareMathAlphabet{\mathsfit}{\encodingdefault}{\sfdefault}{m}{sl}
\SetMathAlphabet{\mathsfit}{bold}{\encodingdefault}{\sfdefault}{bx}{n}

\def\sH{{\mathbb{H}}}

\def\sX{{\mathbb{X}}}
\def\sY{{\mathbb{Y}}}

\newcommand{\R}{\mathbb{R}}



%% file: papel/import.tex
\usepackage{graphicx}
\usepackage{pifont} 
\newcommand{\greencheck}{\textcolor{green!60!black}{\ding{51}}}
\newcommand{\redx}{\textcolor{red}{\ding{55}}}
\usepackage{booktabs} 

%% file: papel/abstract.tex
\begin{abstract}
Self-supervised learning in healthcare has largely relied on invariance-based objectives, which maximize similarity between different views of the same patient. While effective for static anatomy, this paradigm is fundamentally misaligned with clinical diagnosis, as it mathematically compels the model to suppress the transient pathological changes it is intended to detect. We propose a shift towards Action-Conditioned World Models that learn to simulate the dynamics of disease progression, or \emph{Event-Conditioned}. Adapting the LeJEPA framework to physiological time-series, we define pathology not as a static label, but as a transition vector acting on a patient's latent state. By predicting the future electrophysiological state of the heart given a disease onset, our model explicitly disentangles stable anatomical features from dynamic pathological forces. Evaluated on the MIMIC-IV-ECG dataset, our approach outperforms fully supervised baselines on the critical triage task. Crucially, we demonstrate superior sample efficiency: in low-resource regimes, our world model outperforms supervised learning by over 0.05 AUROC. These results suggest that modeling biological dynamics provides a dense supervision signal that is far more robust than static classification. Source code is available at our \href{https://github.com/cljosegfer/lesaude-dynamics}{GitHub repository}.
\end{abstract}

%% file: papel/introduction.tex
\section{Introduction}

The primary goal of clinical monitoring is the detection of change. Whether observing a patient in an intensive care unit or tracking chronic conditions over years of outpatient visits, the clinical value of a diagnostic tool lies in its ability to distinguish the stable anatomical baseline of an individual from acute pathological transitions. However, the current landscape of self-supervised learning (SSL) in medical artificial intelligence is governed by a fundamental contradiction: it is built almost entirely on the principle of \textit{invariance}.

In standard SSL frameworks, such as SimCLR or MoCo, and their medical adaptations like PCLR~\citep{diamant2022patient} and MICLe~\citep{azizi2021big}, the objective is to learn representations that remain stable under perturbations. In the context of longitudinal data, this often manifests as \emph{Patient-Invariance}, where a model is trained to maximize the similarity between different recordings of the same individual. While effective for tasks involving static anatomical characterization or patient re-identification, we argue that this paradigm is inherently misaligned with the requirements of acute care. By forcing a model to map a healthy baseline and an acute pathological state (e.g., the onset of a paroxysmal arrhythmia) to the same point in a latent manifold, the encoder is mathematically compelled to discard the transient disease signals as noise or augmentation artifacts~\citep{mehari2022self}. This creates a fundamental information bottleneck: a robust biometric model that is paradoxically blind to the very changes it is meant to monitor.

Furthermore, the standard data augmentations used to provide the \emph{views} for contrastive learning—such as random cropping, jittering, or scaling—often fail when applied to 1D physiological signals like the Electrocardiogram (ECG). Techniques like random cropping can be semantically destructive, potentially removing the P-wave and rendering a model unable to assess the P-R interval~\citep{andersson2023self}, while jittering and scaling can inadvertently mimic serious pathologies~\citep{nonnenmacher2022utilizing}. Consequently, a model trained to be invariant to these transformations may unlearn critical clinical markers essential for diagnosis.

To resolve these limitations, we propose a shift from static invariance to \textit{dynamic causality}. We introduce the Action-Conditioned Cardiac World Model, a framework that formalizes the monitoring problem as a simulation of the heart's electrophysiological state under pathological perturbations. Rather than suppressing longitudinal variations, our model explicitly treats disease onset as an \textit{action}—a translational force that moves the patient's representation through a latent manifold. By utilizing the Joint-Embedding Predictive Architecture (JEPA)~\citep{lecun2022path}, we avoid the computational burden and "hallucination" risks associated with generative pixel-reconstruction models like MeWM~\citep{yang2025medical}. Instead, we predict future states directly in a geometrically regularized latent space.

This approach allows the model to disentangle the stable anatomical substrate (Patient Identity) from the evolving disease physics. Our experiments on the MIMIC-IV-ECG dataset demonstrate that this dynamical systems perspective achieves parity with fully supervised models on longitudinal monitoring while significantly outperforming them on the critical task of triage for new patients. Most importantly, we show that dynamics modeling provides a denser supervision signal than classification, maintaining high performance in low-data regimes where supervised baselines suffer catastrophic degradation.

Our contributions are as follows:
\begin{itemize}
    \item We propose an \textbf{Action-Conditioned World Model} for medical diagnosis, recontextualizing disease onset as a translational action variable.
    \item We demonstrate that \textbf{predictive dynamics serve as a superior supervision signal} to standard classification, providing a 0.05 AUROC improvement in low-resource (10\%) data regimes.
    \item We show that our model \textbf{overcomes the limitations of standard data augmentations} by using longitudinal transitions as the primary source of supervision, resulting in more robust representations for clinical monitoring.
\end{itemize}

%% file: papel/related.tex
\section{Related Work}

\paragraph{Invariance-based Self-Supervised Learning.} The dominant paradigm in medical representation learning has centered on the principle of invariance. Approaches such as Patient Contrastive Learning of Representations (PCLR)~\citep{diamant2022patient} maximize similarity between different recordings of the same individual to learn robust anatomical features. While successful for static tasks like sex classification or age regression, this patient-invariance creates a fundamental information bottleneck for clinical monitoring. As demonstrated by the observed \emph{AFib Gap}, forcing a model to map a healthy baseline and an acute pathological state (e.g., paroxysmal Atrial Fibrillation) to the same latent point compels the encoder to discard the very disease signals it is intended to detect. Similar limitations are observed in Multi-Instance Contrastive Learning (MICLe)~\citep{azizi2021big}, which excels in static domains like dermatology but collapses when applied to transient pathologies. Furthermore, efforts to remove subject variability in EEG~\citep{salsabilian2022subject} often result in a loss of sensitivity to personalized anomalies, such as pre-ictal states in epilepsy. Our work diverges from this paradigm by treating patient identity not as a nuisance to be invariant to, but as a stable substrate that must be disentangled from dynamic pathological transitions.

\paragraph{Medical World Models.} The concept of "World Models"—internal simulations that predict the consequences of actions—has recently transitioned from robotics to healthcare. Generative World Models, such as MeWM~\citep{yang2025medical}, simulate tumor evolution conditioned on treatment plans but rely on high-fidelity pixel reconstruction, which is computationally expensive and prone to hallucinations that lack biological reality. More recent efforts like SurgWM~\citep{shah2026learning} utilize latent dynamics to \emph{imagine} surgical tissue deformation based on tool movement, demonstrating the utility of predicting in embedding space. In the context of static imaging, CheXWorld~\citep{yue2025chexworld} treats spatial exploration as a dynamic process, using a JEPA framework to disentangle anatomy from domain variation. Our approach extends this logic to the temporal domain of ECG, recontextualizing disease onset not as a static label, but as a translational \emph{action} on the patient’s latent identity.

\paragraph{Latent Dynamics and JEPA in Time-Series.} Our architecture builds upon the Joint-Embedding Predictive Architecture (JEPA)~\citep{lecun2022path}, which avoids the pitfalls of both generative reconstruction and contrastive invariance. While Masked Autoencoders (MAE) approaches like \citet{zhou2025multi} focus on high-frequency texture and sensor noise~\citep{zhang2024self}, JEPA-style predictive objectives~\citep{ren2022contrastive} capture semantic causal structures by predicting future latent states. JETS~\citep{xie2025jets} represents a primary application of JEPA to physiological time-series; however, it remains purely autoregressive (passive). Our model addresses this action gap by injecting a pathology transition vector, upgrading the model from a passive forecaster to an active simulator of cardiac evolution. This transition is further supported by literature on time-irreversibility~\citep{agrawal2022leveraging}, which argues that because disease progression is entropic and directed, models must respect the "arrow of time" rather than treating longitudinal pairs as interchangeable views.

\paragraph{Disease Progression as a Dynamical System.} Modeling pathology as a trajectory through a latent manifold aligns with recent advances in Neural Ordinary Differential Equations (NODEs). LNODE~\citep{wen2025lnode} demonstrates the power of disentangling subject-specific parameters (identity) from cohort-shared parameters (disease physics) to improve forecasting. While previous longitudinal SSL methods (LSSL)~\citep{zhao2021longitudinal} regularize latent space based on temporal smoothness, they often struggle with the discontinuous jumps caused by acute events like myocardial infarction. By conditioning on an explicit action vector, our framework linearizes these complex differential equations, allowing for the simulation of both gradual progression and sudden pathological onsets. This bridges the gap between self-supervised representation learning and causal counterfactual regression~\citep{el2024causal}, enabling the model to answer "what-if" questions regarding a patient’s cardiac future.

\paragraph{Limitations of Data Augmentation in ECG.} Finally, our move toward predictive dynamics is motivated by the failure of standard SSL augmentations for periodic signals. Common techniques like random cropping can be semantically destructive in ECG, potentially removing the P-wave and rendering the model unable to assess the P-R interval~\citep{andersson2023self}. As noted in recent critiques of ECG augmentation \citep{mehari2022self, nonnenmacher2022utilizing}, standard transformations like jittering and scaling risk introducing invariances to features that are morphologically similar to acute pathologies. \citet{mehari2022self} demonstrated that such artificial views can lead to significantly reduced robustness compared to supervised baselines, effectively causing the model to unlearn critical clinical markers. By leveraging longitudinal transitions as the primary supervision signal, our model avoids the need for ad-hoc heuristics that risk unlearning critical clinical features.

%% file: papel/method.tex
\section{Method: Action-Conditioned Cardiac World Models}
\begin{figure}[ht]
\begin{center}
\centerline{\includegraphics[width=\columnwidth]{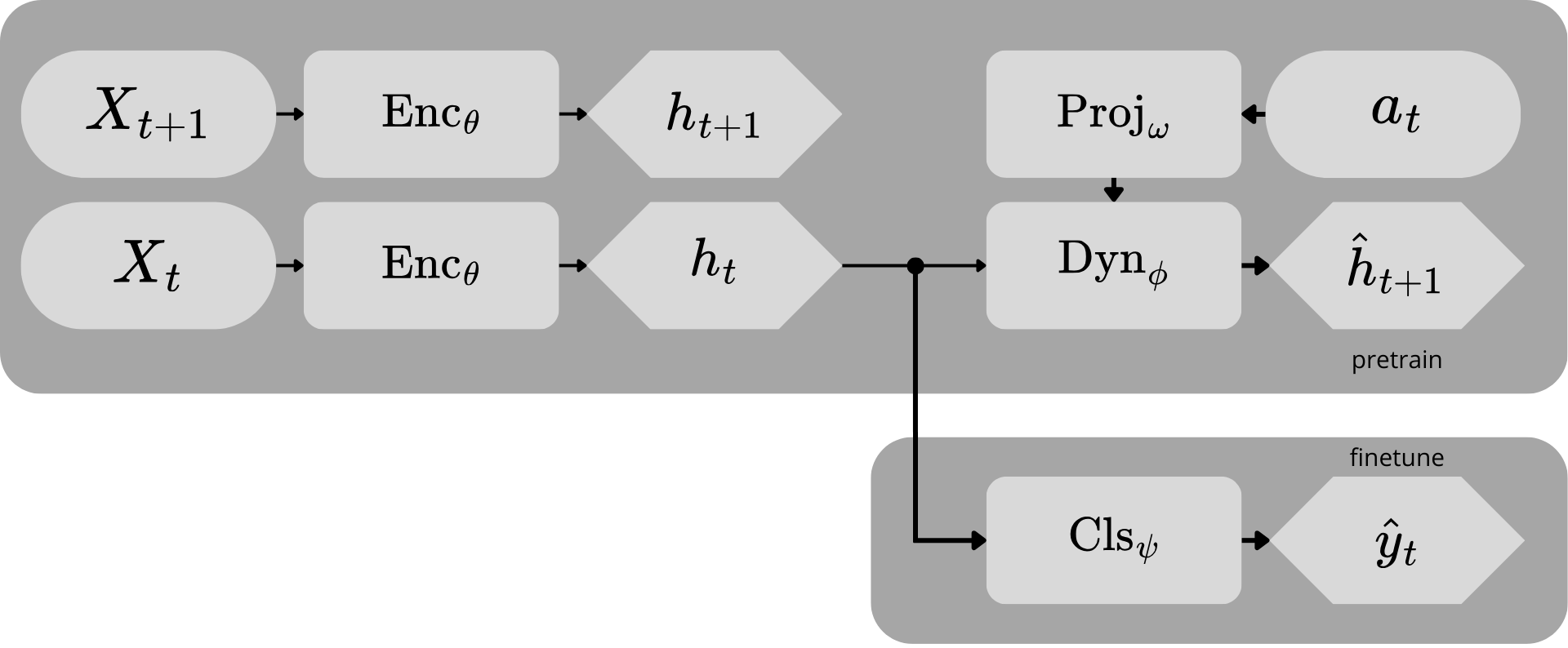}}
\end{center}
\caption{Proposed Action-Conditioned World Model Architecture. The framework consists of two stages. Top (Pre-training): the encoder $\text{Enc}_\theta$ maps longitudinal ECG pairs ($\mX_t, \mX_{t+1}$) to latent representations ($\vh_t, \vh_{t+1}$). A dynamics network $\text{Dyn}_\phi$ predicts the future latent state $\hat{\vh}_{t+1}$ conditioned on the current state $\vh_t$ and the pathology transition vector $\va_t = \vy_{t+1}-\vy_t$, processed by projector $\text{Proj}_\omega$. Bottom (Fine-tuning): the learned representation $\vh_t$ is passed to a classifier $\text{Cls}_\psi$ to predict multi-label diagnoses $\hat{\vy}_t$.}
\label{fig:architecture}
\end{figure}

We formalize the cardiac monitoring problem as learning a world model that simulates the evolution of the heart’s electrophysiological state under pathological perturbations. Unlike standard SSL which enforces invariance to augmentation, our objective is to learn a predictive latent space where disease onset acts as a translational force on the patient’s anatomical representation, leveraging patient identity information. The architecture is schematized as in Figure~\ref{fig:architecture}.

\paragraph{Problem Formulation.}
Let $\sX$ be the space of 12-lead ECG signals $\mX \in \R^{12 \times T}$ and $\sY$ be the associated set of diagnosis labels $\vy \in \{0,1\}^C$. In longitudinal datasets (e.g., MIMIC-IV), we observe a sequence of states $({\mX}_t, {\vy}_t)$ for a specific patient.

We define the action $\va_t$ not as the static label, but as the transition vector describing the pathological change between time steps:
$$ \va_t = \vy_{t+1} - \vy_t \in \{-1, 0, 1\}^C $$
Here, $\va_{t,i}=1$ denotes disease onset, $-1$ denotes resolution, and $0$ implies stability. The goal of the World Model is to learn an encoder $\text{Enc}_\theta: \sX \rightarrow \sH$ and a predictor $\text{Dyn}_\phi$ such that the future latent state can be estimated from the current state and the applied clinical \emph{action}, after a projection via $\text{Proj}_\omega$:
$ \hat{\vh}_{t+1} = \text{Dyn}_\phi(\text{Enc}_\theta(\mX_t), \text{Proj}_\omega(\va_t)) $.
This formulation forces the encoder $\text{Enc}_\theta$ to disentangle Patient Identity (invariant features required to ground the prediction) from Disease State (dynamic features modified by $\va_t$).

\begin{figure}[htb]
\begin{center}
\centerline{\includegraphics[width=\columnwidth]{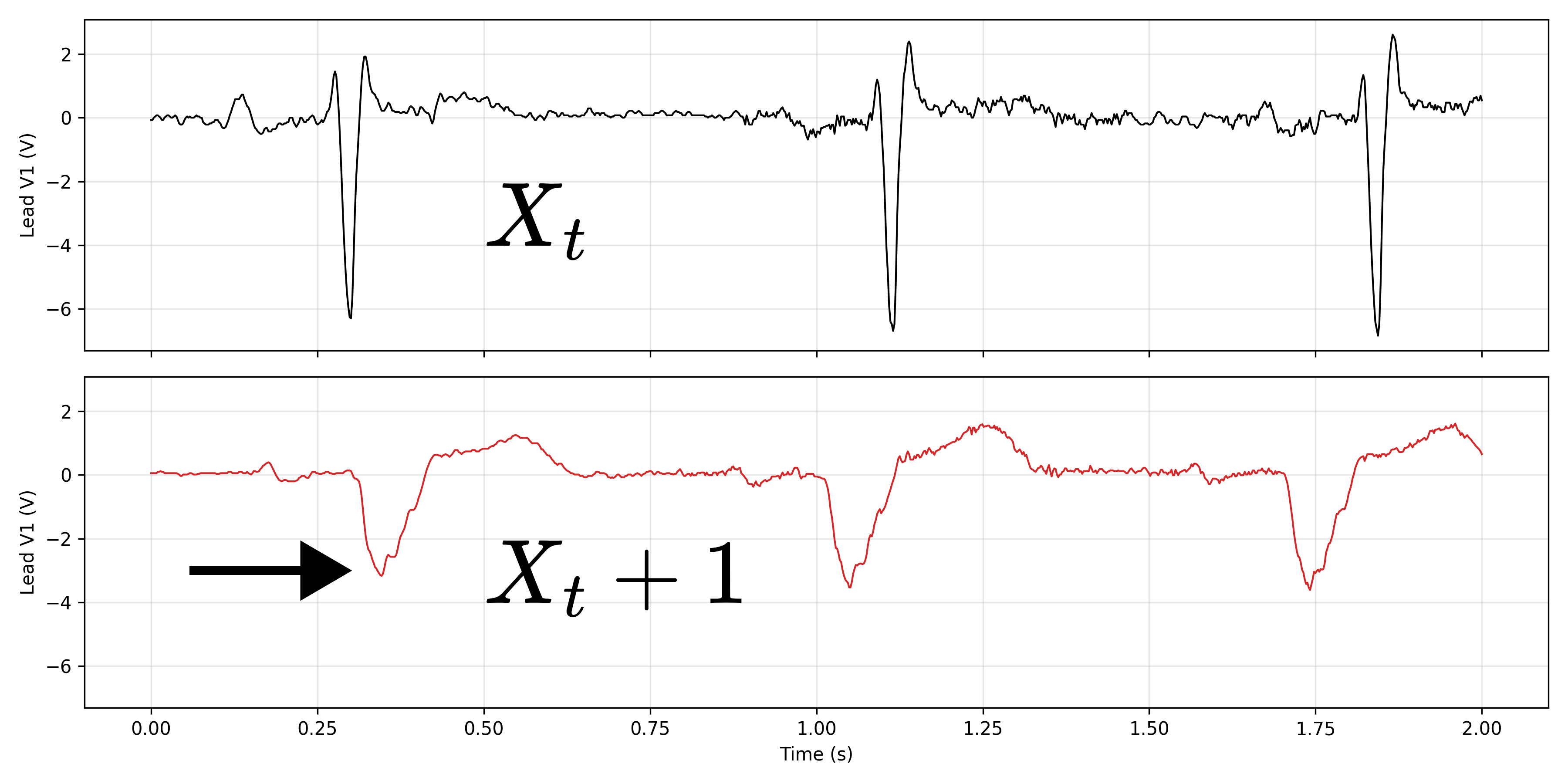}}
\end{center}
\caption{
Visualization of a Pure Pathological Transition. This longitudinal pair illustrates a transition where the action vector $\va_t$ is sparse, representing the onset of a single pathology (I44). The baseline signal (top) shows a normal ventricular activation, while the subsequent recording (bottom) exhibits the classic morphological hallmark of I44, a widened QRS complex (black arrow). By conditioning on this transition vector, the world model is trained to simulate the transformation from a healthy baseline to this pathological state within the latent manifold.}
\label{fig:transition0}
\end{figure}

\begin{figure}[htb]
\begin{center}
\centerline{\includegraphics[width=\columnwidth]{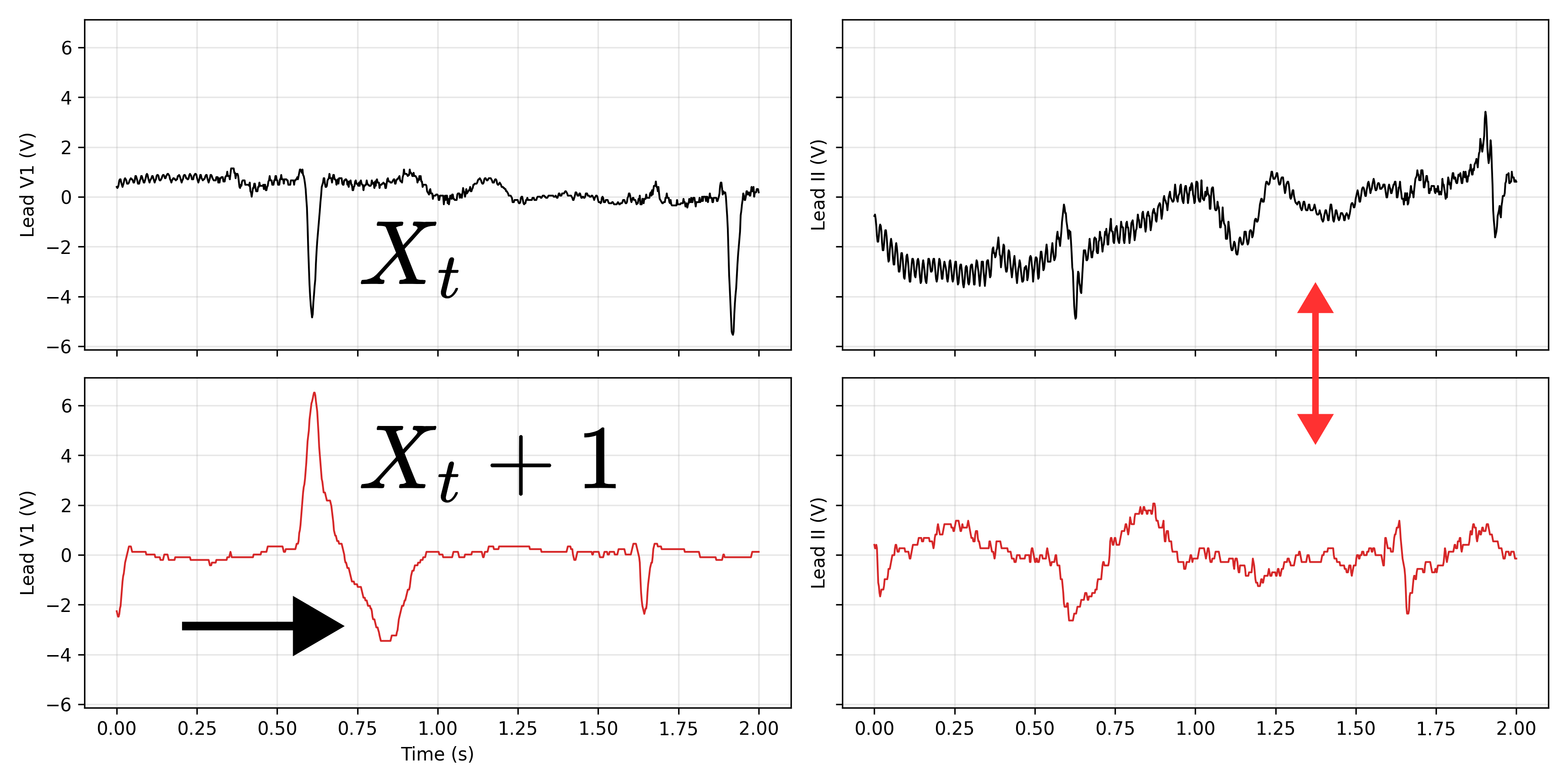}}
\end{center}
\caption{Disentanglement of Acute Pathology from Stable Anatomy. In this more complex clinical scenario, both the baseline and the future state share a stable underlying condition (I48), identified by the persistent flutter waves (red arrow). However, the transition vector $\va_t$ again signals the acute onset of I44 (wide QRS complex, black arrow). This demonstrates the model’s objective, it must remain invariant to the action-transparent features (flutter waves) while accurately predicting the pathological ones (wide QRS), effectively distinguishing the acute event from the chronic background.}
\label{fig:transition1}
\end{figure}

\paragraph{Dynamics Visualization.}
To illustrate the transition vector's role in grounding the latent space, we visualize specific longitudinal ECG pairs ($\mX_t,\mX_{t+1}$). In Figure~\ref{fig:transition0}, the action $\va_t$ is active only for label I44, which is morphologically represented by the wide QRS complex (black arrow). In Figure~\ref{fig:transition1}, while $\va_t$ remains still active for I44, both signals share a common stable label in $\vy$ (I48), consequently, this component is transparent in the action vector $\va_{t,\text{I48}}=\vy_{t+1,\text{I48}}-\vy_{t,\text{I48}}=0$, distinct from the flutter waves (red arrow) visible in the traces. Critically, these examples illustrate the disentanglement objective, the model is trained to linearize disease progression, isolating the transition vector as an independent force acting upon a stable, patient-specific substrate.

\paragraph{Architecture.}
We adopt the JEPA framework, adapted for 1D physiological signals. The system consists of an encoder $\text{Enc}_\theta$ implemented as an xResNet1d50, a 1D-adapted ResNet architecture optimized for time-series following previous ECG classification literature~\citep{strodthoff2024prospects}, which maps the raw waveform $\mX_t$ to a latent representation $\vh_t$. Simultaneously, the sparse difference vector $a_t$ is projected into a continuous action embedding via a multi-layer perceptron (MLP) as a projector $\text{Proj}_\omega$. The Latent Dynamics Predictor $\text{Dyn}_\phi$, a residual MLP, then estimates the future representation $\hat{\vh}_{t+1} = \text{Dyn}_\phi(\text{Enc}_\theta(\mX_t), \text{Proj}_\omega(\va_t))$ based on the current state and given the action. Note that unlike generative models, we predict in latent space, avoiding the complexity of pixel-level reconstruction.

\paragraph{Regularization via SIGReg.}
A critical failure mode in predictive SSL is representation collapse (mapping all inputs to a constant vector). While methods like VICReg~\citep{bardes2021vicreg} prevent this via heuristic variance-covariance constraints, they introduce brittle hyperparameters that are difficult to tune. We instead employ Sketched Isotropic Gaussian Regularization (SIGReg) from the LeJEPA framework. SIGReg enforces that the distribution of embeddings $\sH$ follows a standard Isotropic Gaussian $\mathcal{N}(\vh;\mathbf{0}, \mI)$ by minimizing the distance between their empirical characteristic functions projected onto random 1D slices.

The total training objective is:
$$ \mathcal{L} = (1-\lambda) \underbrace{\| \hat{\vh}_{t+1} - \vh_{t+1} \|^2}_{\text{Prediction (Dynamics)}} + \lambda \underbrace{(\text{SIGReg}(\mH_t) + \text{SIGReg}(\mH_{t+1}))}_{\text{Geometry Constraint}} $$

By enforcing an isotropic Gaussian geometry, SIGReg ensures the latent space is maximally entropic (preventing collapse) and rotationally symmetric, which is theoretically optimal for linear separability in downstream tasks. This allows our model to learn rich, manipulable representations of cardiac dynamics without requiring the extensive grid searches associated with contrastive methods.

\paragraph{Downstream Task.}
To assess whether the learned world model captures clinically relevant features, we evaluate the encoder $\text{Enc}_\theta$ on downstream multi-label classification tasks under two distinct regimes: Linear Probing (Geometry Check); and, Finetuning (Adaptability Check). For the former we freeze the encoder and train a linear head $\text{Cls}_\psi$ on top of $\vh_t$. This tests if the latent space is linearly separable by disease, validating that the SIGReg geometry constraint successfully organized the pathological states. Then, in finetuning, we unfreeze the encoder and update all weights using the supervised classification loss. This tests if the pre-trained initialization places the model in a better optimization basin than random initialization, particularly in low-data regimes.

For both protocols, we utilize the Asymmetric Loss to handle the extreme class imbalance (e.g., Hypertension vs. Rare Arrhythmias) inherent in the dataset.

%% file: papel/experiments.tex
\section{Experiments and Discussion}

We evaluate our approach on the MIMIC-IV-ECG dataset~\citep{gow2023mimic} from PhysioNet~\citep{goldberger2000physiobank}, containing $\sim$800k records from $\sim$160k patients. We utilize the official patient-stratified folds~\citep{strodthoff2024mimic} to ensure zero identity leakage between splits\footnote{See Appendix~\ref{sec:dataset} for a complete dataset characterization.}. We compare our LeJEPA-based approach against a fully supervised model trained from scratch and a Naive Patient-Based SSL baseline. We implement the latter by mapping $\mathbf{h}_t \to \mathbf{h}_{t+1}$ without action conditioning, effectively treating longitudinal exams as positive pairs for standard invariance learning.

\paragraph{Generalization and Robustness.} We first evaluate performance on the full dataset under two distinct clinical scenarios: Triage, evaluating only on the first ECG of a patient's visit; Monitoring, which evaluates on all ECGs but captures the temporal redundancy inherent in multi-record hospital stays. Results are detailed in Table~\ref{tab:results}.

\begin{table*}[t]
\caption{Macro-AUROC Performance on MIMIC-IV-ECG. We compare our Action-Conditioned Dynamics model against a supervised baseline and a naive patient-invariance SSL baseline. We report results on the Triage task and the Monitoring task. 95\% confidence intervals from 1,000 bootstrap iterations are shown in brackets.}
\label{tab:results}
\centering
\resizebox{\textwidth}{!}{
\begin{tabular}{@{}l l cc c cc@{}}
\toprule
\multicolumn{2}{c}{\textbf{Data Usage}} & \multicolumn{2}{c}{\textbf{}} & \multicolumn{2}{c}{\textbf{Macro-AUROC}} \\
\cmidrule(r){1-2} \cmidrule(lr){3-4} \cmidrule(l){5-6}
\textbf{Labels} ($\vy$) & \textbf{Patient ID} & \textbf{Method} & \textbf{Eval. Protocol} & \textbf{Triage (First ECG)} & \textbf{Monitoring (All)} \\
\midrule

\greencheck & \redx & Supervised & End-to-End & 0.735 \small{[0.727 -- 0.751]} & 0.786 \small{[0.776 -- 0.797]} \\

\addlinespace

\redx & \greencheck & Naive SSL (Invariance) & Linear Probe & 0.666 \small{[0.656 -- 0.682]} & 0.698 \small{[0.689 -- 0.709]} \\
 & & & Finetuned & 0.679 \small{[0.668 -- 0.697]} & 0.726 \small{[0.715 -- 0.737]} \\

\addlinespace

\greencheck & \greencheck & Ours (Latent Dynamics) & Linear Probe & 0.725 \small{[0.714 -- 0.742]} & 0.744 \small{[0.734 -- 0.755]} \\
 & & & Finetuned & 0.742 \small{[0.733 -- 0.756]} & 0.786 \small{[0.777 -- 0.794]} \\

\bottomrule
\end{tabular}
}
\end{table*}

While both methods achieve parity on the easier Monitoring task, our Dynamics model slightly outperforms the Supervised baseline on the critical Triage task. We hypothesize that the dynamics learning—which requires predicting the transformation of the signal—forces the encoder to disentangle anatomical features from pathological changes, resulting in robust representations. The Naive SSL baseline did not yield good downstream performance for this task, confirming that patient invariance alone is insufficient for diagnosis.

\paragraph{Data Efficiency.} To test the hypothesis that dynamics modeling provides a denser supervision signal than classification, we simulate low-resource environments by restricting the training data to 10\% ($\sim$4k patients) and 1\% ($\sim$400 patients) of the dataset. Crucially, unlike standard SSL, we also pretrain with this restriction as our approach requires the label.

\begin{figure}[ht]
\begin{center}
\centerline{\includegraphics[width=\columnwidth]{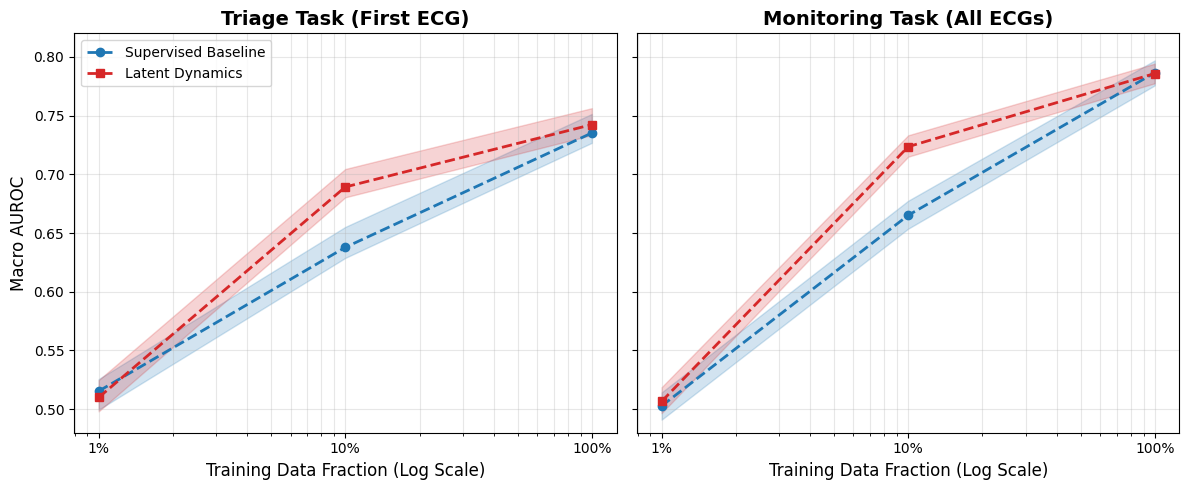}}
\end{center}
\caption{Macro-AUROC vs. Data Fraction. Results are shown for Triage (First ECG) and Monitoring (All ECGs) tasks. The Latent Dynamics model demonstrates significantly higher sample efficiency in the 10\% regime ($p < 0.05$, non-overlapping intervals) compared to the fully supervised baseline.}
\label{fig:lowdata}
\end{figure}

As illustrated in Figure~\ref{fig:lowdata}, the Supervised model suffers catastrophic degradation in the 10\% regime ($0.74 \rightarrow 0.64$ AUROC), owing to the high parameter-to-sample ratio. Conversely, our Dynamics model maintains robustness ($0.74 \rightarrow 0.69$ AUROC), significantly outperforming the baseline ($p < 0.05$). This suggests that predicting a high-dimensional future latent state $\vh_{t+1}$, given the action, provides a richer gradient signal than categorical labels, enabling complex feature learning even when data is scarce. At 1\% data, we observe a capability collapse where both models degrade to random guessing ($\sim$0.50 AUROC). This establishes a lower bound for the xResNet1d50 architecture on this modality, indicating that below this threshold, model capacity exceeds the information content of the subset regardless of the objective.

\paragraph{Failure-mode.}
To investigate this performance degradation at the 1\% data regime, we conduct a post-mortem analysis comparing full-parameter finetuning against linear probing on frozen representations. Results are detailed in Table~\ref{tab:results001}. We observe a striking divergence: while full-parameter finetuning collapses to random guessing, the linear probe maintains a significantly higher signal. 

This suggests that the pretraining successfully captures cardiac dynamics and organizes a meaningful latent space even at this extreme low-data scale. However, the standard supervised finetuning objective appears to introduce sufficient gradient noise to overwrite these learned features when labels are sparse. This finding confirms that the collapse is not a failure of the world model's representation learning, but rather a failure of the task-specific adaptation protocol, highlighting the need for more robust distillation methods.

\begin{table*}[t]
\caption{Low-Data Regime Analysis (1\% Data). We compare the Supervised baseline against our World Model using two evaluation protocols. The gap between Linear Probe and Finetuned results indicates that the world model retains clinical features that are destroyed during full-parameter finetuning.}
\label{tab:results001}
\centering
\begin{tabular}{@{}l l c cc@{}}
\toprule
\multicolumn{2}{c}{\textbf{}} & \multicolumn{2}{c}{\textbf{Macro-AUROC}} \\
\cmidrule(lr){1-2} \cmidrule(l){3-4}
\textbf{Method} & \textbf{Eval. Protocol} & \textbf{Triage (First ECG)} & \textbf{Monitoring (All)} \\
\midrule

Supervised & End-to-End & 0.515 \small{[0.500 -- 0.526]} & 0.503 \small{[0.491 -- 0.514]} \\

\addlinespace

Ours (Latent Dynamics) & Linear Probe & 0.563 \small{[0.551 -- 0.575]} & 0.575 \small{[0.560 -- 0.591]} \\
& Finetuned & 0.510 \small{[0.498 -- 0.525]} & 0.507 \small{[0.497 -- 0.519]} \\

\bottomrule
\end{tabular}
\end{table*}

%% file: papel/conclusion.tex
\section{Conclusion and Future Work}

We presented an Action-Conditioned World Model for cardiac monitoring that learns to simulate the electrophysiological evolution of the heart. By decoupling patient identity from pathological transitions via the LeJEPA architecture, our method achieves parity with fully supervised baselines on longitudinal monitoring while slightly outperforming them on the critical Triage task ($0.742$ vs $0.735$ AUROC). Most notably, in low-resource regimes (10\% data), our dynamics objective serves as a potent regularizer, improving performance by over 0.05 points with confidence, showing evidence that the dense signal of predicting future states, given the label-action, is more sample-efficient than static classification. For future work, we envision three key directions to scale this framework:

\paragraph{Data efficiency and the limits of label-as-action.}
Since our predictor approximates a reversible dynamical system, we can generate synthetic training
data by reversing transitions ($h_{t+1}, -a_t \rightarrow \hat{h}_t$) or chaining multiple
pathology vectors, potentially unlocking performance in the 1\% data regime where current models
collapse. However, the validity of reversal is not uniform across pathologies and opens a deeper
question about the representability of $a_t$ itself. Disease progression is fundamentally entropic
and temporally directed~\citep{agrawal2022leveraging}. For example, the resolution of a paroxysmal arrhythmia (I48)
is a plausible clinical trajectory and its reversal, onset from a clean baseline, is a
symmetric and informative counterfactual. By contrast, reversal is physically inadmissible for
irreversible conditions such as myocardial infarction (I21), where the
\emph{arrow of time} is strictly one-directional. Crucially, this asymmetry reveals a broader
limitation of the binary action space $\{-1, 0, 1\}^C$, it encodes only the \textit{fact} of a
transition, not its \textit{clinical magnitude}. An acute
arrhythmia onset and a gradual chronic drift are aliased to the same $a_{t,i} = 1$ regardless of
their fundamentally different dynamics.

This aliasing is most consequential for the chronic conditions that dominate our dataset.
Conditions such as Hypertension (I10) and Ischemic Heart Disease (I25), the two most frequent
labels in MIMIC-IV-ECG, progress over years rather than timesteps, meaning the transition vector
will be near-zero for the vast majority of their sequential pairs. The world model, therefore, might be
well-regularized for discrete acute events but receives a weak and infrequent training signal for
the slow drift of chronic substrates. Furthermore, in real clinical data the precise onset of even
acute pathologies is often temporally ambiguous. ICD-10 codes may be assigned retrospectively,
and the physiological change captured in the waveform may precede or lag the recorded label. We
note that MIMIC-IV-ECG mitigates this by pairing each waveform with codes assigned at the time of
acquisition, and that the prevalence of sparse single-label transitions (refer to Appendix~\ref{sec:dataset}) is consistent with discrete clinical events
rather than noisy multi-label drift, but the ambiguity cannot be fully eliminated within a binary
action space.

Both limitations might converge to a single resolution. Future iterations can replace the one-hot
transition vector with a learned continuous Action Embedding ($a_t \in \mathbb{R}^d$). A dense
embedding space can represent gradations of onset, encode temporal asymmetry by construction, and
assign geometrically proximal vectors to clinically related conditions, allowing rare or
slowly-evolving diseases to borrow statistical strength from common ones. Conditioning the
dynamics predictor additionally on the \textit{elapsed time} $\Delta t$ between recordings would
further disentangle the two regimes: a one-day gap where stability is expected versus a one-year
gap where chronic drift is plausible. This extension aligns naturally with Neural ODE
formulations~\citep{wen2025lnode}, where the latent trajectory is a continuous function of time,
enabling the same framework to capture both the sharp electrophysiological signatures of acute
events and the slow, irreversible drift of chronic disease substrates without requiring separate
model families for each regime.

\paragraph{Label sparsity.} We currently restrict labels to Chapter IX (Circulatory) truncated to 2 digits. The learned Action Embedding described above scales this naturally to the full ICD-10 ontology: unlike supervised heads which require dense samples per class to converge, a dynamics model shares the transformation physics across the embedding space, allowing rare diseases to borrow statistical strength from common ones without the instability typical of high-dimensional classification heads.

\paragraph{Counterfactual Simulation.} Leveraging the geometry enforced by SIGReg, we aim to deploy the model as a patient-specific simulator. By applying a disease vector to a healthy patient's embedding ($\hat{h}_{sick} = \text{Dyn}_\phi(h_{healthy}, a_{disease})$), we can generate counterfactual trajectories---answering \textit{``What would this specific patient look like if they developed Atrial Fibrillation?''}---offering a powerful tool for interpretability. <- contextualized inference

%% file: papel/appendix.tex
\appendix

\section{Dataset Characterization}
\label{sec:dataset}
We evaluate our model using the MIMIC-IV-ECG diagnostic matched subset, a large-scale longitudinal repository of clinical 12-lead ECG waveforms. This appendix provides a detailed statistical characterization of the dataset, the label processing workflow, and the resulting distribution of pathological transitions (actions).

\begin{figure}[ht]
\begin{center}
\centerline{\includegraphics[width=\columnwidth]{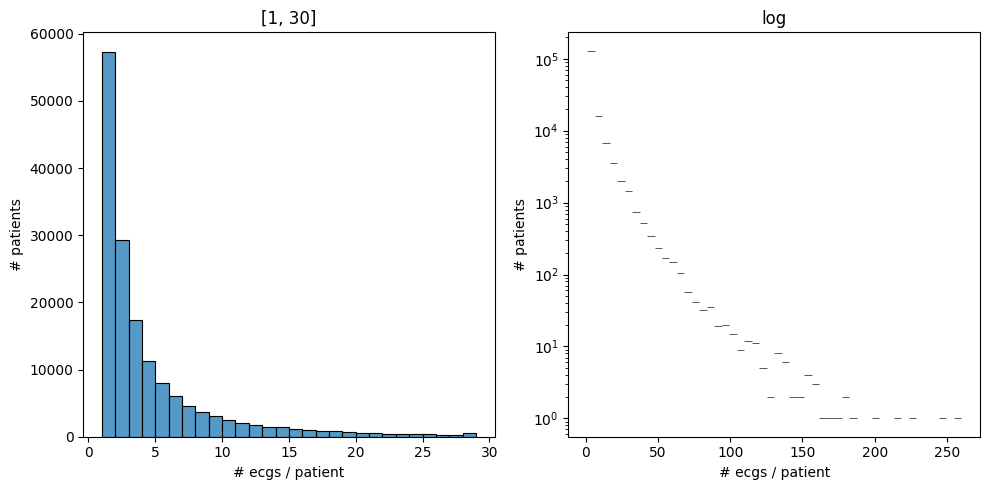}}
\end{center}
\caption{Distribution of longitudinal ECG recordings per patient. The left panel shows a histogram for the bulk of the cohort (1 to 30 recordings per patient), demonstrating that while many patients are singletons, the majority contribute multiple observations. The right panel provides a full-scale view on a logarithmic y-axis, highlighting the dataset's significant long tail.}
\label{fig:hist}
\end{figure}

\paragraph{Data Scale and Longitudinality.}
The dataset contains a total of $800,035$ ECG records associated with $161,352$ unique patients. A defining feature of this dataset is its longitudinal nature, which is essential for training world models. While $57,344$ patients are represented by only a single recording (singletons), the majority of the cohort ($104,008$ patients) have two or more recordings. On average, each patient has approximately $4.96$ ECGs (median: $2.0$; maximum: $260$). See Figure~\ref{fig:hist}.

By extracting consecutive recordings for each patient, we identified $638,683$ unique sequential pairs $(X_t, X_{t+1})$. These pairs serve as the primary unit of observation for our action-conditioned pre-training, where the model learns the transition dynamics from $t$ to $t+1$.

\paragraph{Label Processing and Cardiac State Definition.}
Each ECG in the dataset is annotated with ICD-10 codes, with an average of $8.15$ codes per record across $15,197$ unique medical conditions. To focus the world model on cardiac electrophysiology, we filtered these labels for ICD-10 Chapter IX (Diseases of the Circulatory System), identified by the prefix `I`. 

Following standard clinical grouping, we truncated these codes to their first three digits (e.g., grouping `I48.91` and `I48.1` into the broader category `I48` for Atrial Fibrillation/Flutter). This reduction resulted in an action space of $76$ distinct cardiac clusters. Under this specialized label set, the average number of cardiac-specific codes per ECG is $1.45$. The most frequent clusters in the dataset include I10 (Hypertension), I25 (Ischemic Heart Disease), I48 (Atrial Fibrillation/Flutter), and I50 (Heart Failure).

\begin{figure}[ht]
\begin{center}
\centerline{\includegraphics[width=\columnwidth]{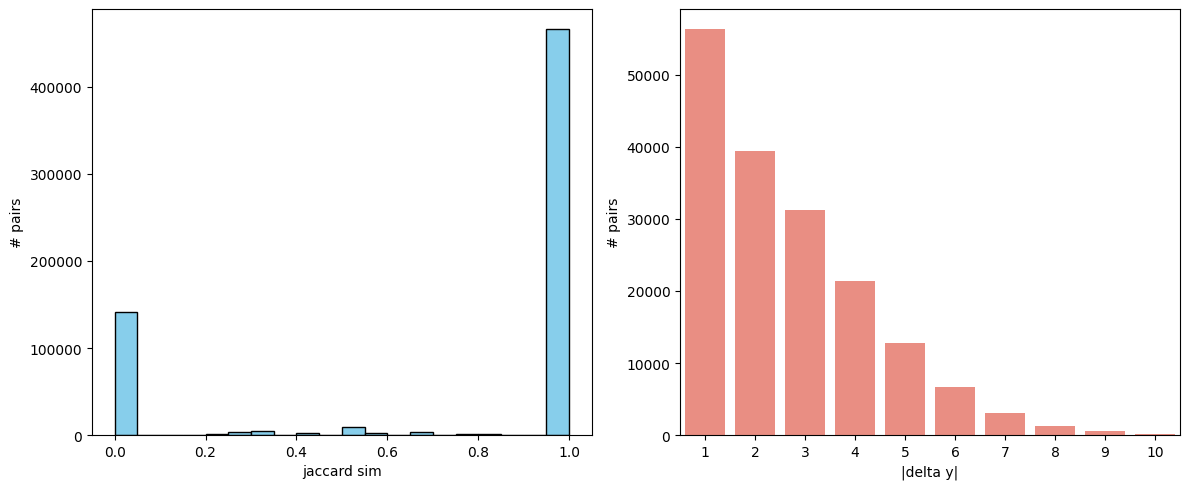}}
\end{center}
\caption{Characterization of cardiac label transitions. The left panel shows the distribution of Jaccard similarity between consecutive label sets $(y_t, y_{t+1})$. The prominent peak at 1.0 represents stable pairs with identical cardiac states, while the remainder identifies the changed pairs in the dataset. The right panel details the magnitude of these actions ($L_1$ norm of the transition vector) for all changed pairs. The prevalence of sparse transitions—specifically single-label changes ($32.6\%$)—supports the formulation of disease onset as a localized translational force in the latent manifold.}
\label{fig:actions}
\end{figure}

\paragraph{Characterizing Pathological Actions.}
We define an "Action" as the transition vector $\va_t = \vy_{t+1} - \vy_t$. Across the $638,683$ sequential pairs, we observed the following distribution of dynamics, also see Figure~\ref{fig:actions}:
\begin{itemize}
    \item \textbf{Stable Pairs (Identity):} $465,842$ pairs ($72.9\%$) showed no change in cardiac ICD-10 clusters between time steps. These pairs provide the "identity" signal, forcing the encoder to learn representations that remain invariant when no action occurs.
    \item \textbf{Changed Pairs (Action):} $172,841$ pairs ($27.1\%$) involve a change in the cardiac state. These transitions are contributed by $55,808$ unique patients, with a mean of $3.10$ "changed" pairs per contributing patient.
\end{itemize}


The "magnitude" of these actions (defined as the $L_1$ norm of the transition vector) is typically sparse. Among the changed pairs, the most common magnitude is a single label transition (adding or removing one disease state), which accounts for $32.6\%$ of all actions. Changes involving $2$ or $3$ labels account for $22.8\%$ and $18.1\%$ respectively. This sparsity supports our formulation of disease onset as a translational force within the latent space, where most clinical events represent specific, localized movements on the latent manifold.


\section{Experimental Setup}
\paragraph{Hardware and Environment.} All experiments were conducted on a single node equipped with an NVIDIA RTX 4090 (24GB VRAM). To mitigate the I/O bottlenecks inherent to processing $\sim$800k small waveform files, we consolidated the dataset into a contiguous HDF5 monolith (stored in Float16 precision) and utilized a persistent RAM caching strategy for the low-data regimes.

\begin{table*}[h]
\caption{Regularization Ablation (Full Dataset). Comparison of the Latent Dynamics model pre-trained with VICReg versus SIGReg. Results are reported for the fine-tuned evaluation protocol. SIGReg outperforms VICReg on both tasks, validating the choice of Isotropic Gaussian geometry over the hypercube constraints of VICReg for physiological signals.}
\label{tab:vicreg}
\centering
\begin{tabular}{@{}l c cc@{}}
\toprule
\multicolumn{1}{c}{\textbf{}} & \multicolumn{2}{c}{\textbf{Macro-AUROC}} \\
\cmidrule(lr){1-1} \cmidrule(l){2-3}
\textbf{Regularization} & \textbf{Triage (First ECG)} & \textbf{Monitoring (All)} \\
\midrule

VICReg & 0.720 \small{[0.710 -- 0.738]} & 0.770 \small{[0.759 -- 0.780]} \\

\addlinespace

SIGReg & 0.742 \small{[0.733 -- 0.756]} & 0.786 \small{[0.777 -- 0.794]} \\

\bottomrule
\end{tabular}
\end{table*}

\paragraph{Architecture and Optimization.} We utilized an xResNet1d50 backbone. For the LeJEPA pre-training, the projector was a 3-layer MLP expanding the 256-dim backbone output to a 256-dim latent space (matching the backbone capacity to avoid information bottlenecks). The predictor was a residual MLP with hidden dimension 512.
Training utilized the AdamW optimizer with a OneCycleLR scheduler. We observed that aggressive learning rates (max LR $1e^{-3}$) were stable for the dynamics task, whereas the supervised baseline required lower rates ($1e^{-4}$) and gradient clipping to prevent divergence. The batch size was set to 512 to maximize GPU throughput, which necessitated careful scaling of the loss terms.


\paragraph{Regularization Ablation.} Prior to adopting SIGReg, we attempted to train the dynamics model using the VICReg objective. As shown in Table~\ref{tab:vicreg}, this approach yielded suboptimal performance compared to SIGReg. We attribute this to VICReg's sensitivity to its three hyperparameters ($\lambda, \mu, \nu$), which typically require extensive grid search to balance. Due to computational constraints, we adopted a heuristic tuning strategy: monitoring the gradient magnitude ratio between the Prediction and Regularization terms during the first epoch. While this allowed us to stabilize SIGReg by simply scaling its single parameter $\lambda$, achieving a similar equilibrium with the multi-term VICReg objective proved intractable on this modality.


%% file: papel/refs.bib
@article{strodthoff2024prospects,
  title={Prospects for artificial intelligence-enhanced electrocardiogram as a unified screening tool for cardiac and non-cardiac conditions: an explorative study in emergency care},
  author={Strodthoff, Nils and Lopez Alcaraz, Juan Miguel and Haverkamp, Wilhelm},
  journal={European Heart Journal-Digital Health},
  volume={5},
  number={4},
  pages={454--460},
  year={2024},
  publisher={Oxford University Press UK}
}

@article{bardes2021vicreg,
  title={Vicreg: Variance-invariance-covariance regularization for self-supervised learning},
  author={Bardes, Adrien and Ponce, Jean and LeCun, Yann},
  journal={arXiv preprint arXiv:2105.04906},
  year={2021}
}

@article{gow2023mimic,
  title={Mimic-iv-ecg: Diagnostic electrocardiogram matched subset},
  author={Gow, Brian and Pollard, Tom and Nathanson, Larry A and Johnson, Alistair and Moody, Benjamin and Fernandes, Chrystinne and Greenbaum, Nathaniel and Waks, Jonathan W and Eslami, Parastou and Carbonati, Tanner and others},
  journal={Type: dataset},
  volume={6},
  pages={13--14},
  year={2023}
}

@article{goldberger2000physiobank,
  title={PhysioBank, PhysioToolkit, and PhysioNet: components of a new research resource for complex physiologic signals},
  author={Goldberger, Ary L and Amaral, Luis AN and Glass, Leon and Hausdorff, Jeffrey M and Ivanov, Plamen Ch and Mark, Roger G and Mietus, Joseph E and Moody, George B and Peng, Chung-Kang and Stanley, H Eugene},
  journal={circulation},
  volume={101},
  number={23},
  pages={e215--e220},
  year={2000},
  publisher={Lippincott Williams \& Wilkins}
}

@article{strodthoff2024mimic,
  title={Mimic-iv-ecg-ext-icd: Diagnostic labels for mimic-iv-ecg (version 1.0. 1)},
  author={Strodthoff, Nils and Lopez Alcaraz, JM and Haverkamp IV, W},
  journal={PhysioNet. RRID: SCR\_007345 https://doi. org/10.13026/hdyc-1h77},
  year={2024}
}

@article{diamant2022patient,
  title={Patient contrastive learning: A performant, expressive, and practical approach to electrocardiogram modeling},
  author={Diamant, Nathaniel and Reinertsen, Erik and Song, Steven and Aguirre, Aaron D and Stultz, Collin M and Batra, Puneet},
  journal={PLoS computational biology},
  volume={18},
  number={2},
  pages={e1009862},
  year={2022},
  publisher={Public Library of Science San Francisco, CA USA}
}

@article{salsabilian2022subject,
  title={Subject-invariant feature learning for mTBI identification using LSTM-based variational autoencoder with adversarial regularization},
  author={Salsabilian, Shiva and Najafizadeh, Laleh},
  journal={Frontiers in Signal Processing},
  volume={2},
  pages={1019253},
  year={2022},
  publisher={Frontiers Media SA}
}

@inproceedings{azizi2021big,
  title={Big self-supervised models advance medical image classification},
  author={Azizi, Shekoofeh and Mustafa, Basil and Ryan, Fiona and Beaver, Zachary and Freyberg, Jan and Deaton, Jonathan and Loh, Aaron and Karthikesalingam, Alan and Kornblith, Simon and Chen, Ting and others},
  booktitle={Proceedings of the IEEE/CVF international conference on computer vision},
  pages={3478--3488},
  year={2021}
}

@article{yang2025medical,
  title={Medical world model: Generative simulation of tumor evolution for treatment planning},
  author={Yang, Yijun and Wang, Zhao-Yang and Liu, Qiuping and Sun, Shuwen and Wang, Kang and Chellappa, Rama and Zhou, Zongwei and Yuille, Alan and Zhu, Lei and Zhang, Yu-Dong and others},
  journal={arXiv preprint arXiv:2506.02327},
  year={2025}
}

@inproceedings{shah2026learning,
  title={Learning Action-Conditioned World Models for Cataract Surgery from Unlabeled Videos},
  author={Shah, Nisarg A and Xia, Mingze and Sikder, Shameema and Vedula, S Swaroop and Patel, Vishal M},
  booktitle={Medical Imaging with Deep Learning},
  year={2026}
}

@inproceedings{yue2025chexworld,
  title={CheXWorld: Exploring Image World Modeling for Radiograph Representation Learning},
  author={Yue, Yang and Wang, Yulin and Tao, Chenxin and Liu, Pan and Song, Shiji and Huang, Gao},
  booktitle={Proceedings of the Computer Vision and Pattern Recognition Conference},
  pages={20778--20788},
  year={2025}
}

@article{zhou2025multi,
  title={Multi-scale Masked Autoencoder for Electrocardiogram Anomaly Detection},
  author={Zhou, Ya and Yang, Yujie and Gan, Jianhuang and Li, Xiangjie and Yuan, Jing and Zhao, Wei},
  journal={arXiv preprint arXiv:2502.05494},
  year={2025}
}

@article{zhang2024self,
  title={Self-supervised learning for time series analysis: Taxonomy, progress, and prospects},
  author={Zhang, Kexin and Wen, Qingsong and Zhang, Chaoli and Cai, Rongyao and Jin, Ming and Liu, Yong and Zhang, James Y and Liang, Yuxuan and Pang, Guansong and Song, Dongjin and others},
  journal={IEEE transactions on pattern analysis and machine intelligence},
  volume={46},
  number={10},
  pages={6775--6794},
  year={2024},
  publisher={IEEE}
}

@article{ren2022contrastive,
  title={A Contrastive Predictive Coding-Based Classification Framework for Healthcare Sensor Data},
  author={Ren, Chaoxu and Sun, Le and Peng, Dandan},
  journal={Journal of Healthcare Engineering},
  volume={2022},
  number={1},
  pages={5649253},
  year={2022},
  publisher={Wiley Online Library}
}

@inproceedings{xie2025jets,
  title={JETS: A Self-Supervised Joint Embedding Time Series Foundation Model for Behavioral Data in Healthcare},
  author={Xie, Erik and Chang, Wyatt and Martinez, Raquel Rodriguez and Ballinger, Brandon},
  booktitle={NeurIPS 2025 Workshop on Learning from Time Series for Health},
year={2025},
}

@inproceedings{agrawal2022leveraging,
  title={Leveraging time irreversibility with order-contrastive pre-training},
  author={Agrawal, Monica N and Lang, Hunter and Offin, Michael and Gazit, Lior and Sontag, David},
  booktitle={International Conference on Artificial Intelligence and Statistics},
  pages={2330--2353},
  year={2022},
  organization={PMLR}
}

@article{lecun2022path,
  title={A path towards autonomous machine intelligence version 0.9. 2, 2022-06-27},
  author={LeCun, Yann},
  journal={Open Review},
  volume={62},
  number={1},
  pages={1--62},
  year={2022}
}

@inproceedings{wen2025lnode,
  title={LNODE: Uncovering the Latent Dynamics of A $\beta$ in Alzheimer’s Disease},
  author={Wen, Zheyu and Biros, George},
  booktitle={International Conference on Medical Image Computing and Computer-Assisted Intervention},
  pages={313--322},
  year={2025},
  organization={Springer}
}

@article{zhao2021longitudinal,
  title={Longitudinal self-supervised learning},
  author={Zhao, Qingyu and Liu, Zixuan and Adeli, Ehsan and Pohl, Kilian M},
  journal={Medical image analysis},
  volume={71},
  pages={102051},
  year={2021},
  publisher={Elsevier}
}

@article{el2024causal,
  title={Causal contrastive learning for counterfactual regression over time},
  author={El Bouchattaoui, Mouad and Tami, Myriam and Lepetit, Benoit and Courn{\`e}de, Paul-Henry},
  journal={Advances in Neural Information Processing Systems},
  volume={37},
  pages={1333--1369},
  year={2024}
}

@article{andersson2023self,
  title={Self-supervised representation learning from electrocardiogram data for medical applications},
  author={Andersson, Matilda},
  journal={Master’s Theses in Mathematical Sciences},
  year={2023}
}

@article{mehari2022self,
  title={Self-supervised representation learning from 12-lead ECG data},
  author={Mehari, Temesgen and Strodthoff, Nils},
  journal={Computers in biology and medicine},
  volume={141},
  pages={105114},
  year={2022},
  publisher={Elsevier}
}

@inproceedings{nonnenmacher2022utilizing,
  title={Utilizing expert features for contrastive learning of time-series representations},
  author={Nonnenmacher, Manuel T and Oldenburg, Lukas and Steinwart, Ingo and Reeb, David},
  booktitle={International Conference on Machine Learning},
  pages={16969--16989},
  year={2022},
  organization={PMLR}
}
